\documentclass[sigconf]{webmedia}
\usepackage{scalefnt}
\AtBeginDocument{%
  \providecommand\BibTeX{{%
    \normalfont B\kern-0.5em{\scshape i\kern-0.25em b}\kern-0.8em\TeX}}}

\begin{document}

\title{Factuality and Transparency Are All RAG Needs! \\ Self-Explaining Contrastive Evidence Re-ranking}
\subtitle{}
%% Of note is the shared affiliation of the first two authors and the
\author{Francielle Vargas}
\email{francielle.vargas@unesp.br}
\author{São Paulo State University}
\email{}
\affiliation{%
  \institution{}
  \streetaddress{}
  \city{}
  \state{}
  \postcode{}
}

\author{Daniel Pedronette}
\affiliation{%
  \institution{São Paulo State University}
  \streetaddress{1 Th{\o}rv{\"a}ld Circle}
  \city{}
  \country{}}
\email{pedronette@unesp.br}

%\author{Fabrício Benevenuto}
%\affiliation{%
%  \institution{UFMG}
%  \city{}
%  \country{fabricio@dcc.ufmg.br}
%}

\renewcommand{\shortauthors}{Vargas 2024}
\begin{abstract}

This extended abstract introduces Self-Explaining Contrastive Evidence Re-Ranking (CER), a novel method that restructures retrieval around factual evidence by fine-tuning embeddings with contrastive learning and generating token-level attribution rationales for each retrieved passage. Hard negatives are automatically selected using a subjectivity-based criterion, forcing the model to pull factual rationales closer while pushing subjective or misleading explanations apart. As a result, the method creates an embedding space explicitly aligned with evidential reasoning. We evaluated our method on clinical trial reports, and initial experimental results show that CER improves retrieval accuracy, mitigates the potential for hallucinations in RAG systems, and provides transparent, evidence-based retrieval that enhances reliability, especially in safety-critical domains.

\end{abstract}

%\keywords{AI for social good, natural language processing, explanability and interpretability, misinformation, hate speech, and online toxicity, fairness,  responsible ai.}


\maketitle

\section{Introduction}

%The proliferation of misinformation and hate speech has become a pressing global issue. These phenomena not only distort public opinion but also exacerbate social tensions, often leading to real-world harm. Although the literature has often presented vague definitions of misinformation, disinformation, and hate speech, according to the United Nations (UN) \cite{Onu2024}, they represent distinct forms of ``harmful speech'' that, despite their differences, share common characteristics and can have significant long-term impacts. 
%According to the United Nations (UN), the dissemination of fake news can serve as a tool for discrimination against individuals or groups, potentially reaching the threshold of incitement \cite{Onu2024}.

Despite recent advances in Retrieval-Augmented Generation (RAG), recent studies have shown that the benefits of RAG can be undermined by noisy retrieval, where passages that truly support a claim are mixed with those that are only topically related, incomplete, or even contradictory, reducing factual correctness and reliability \cite{liu-etal-2023-evaluating,petroni2020contextaffectslanguagemodels}. This limitation is particularly concerning in sensitive domains such as healthcare, where inaccurate or unsupported answers can have severe consequences, for example, compromising patient safety \cite{sun-etal-2025-fact} or leading to legal \cite{Legislation2016} and ethical \cite{chen-etal-2025-llms,cheng-amiri-2025-equalizeir,ghate-etal-2025-biases,kim-etal-2024-discovering} implications. In addition, most RAG systems are not explicitly parameterized to distinguish evidentially valid passages from texts that are semantically proximal lack factual grounding. Consequently, these systems frequently conflate topical similarity with evidential adequacy, resulting in systematic semantic confounding and the absence of retrieval mechanisms optimized for factual correctness. This limitation, combined with insufficient term-level interpretability, directly compromises the factual robustness of downstream generation—an especially critical concern in biomedical and clinical contexts \cite{sun-etal-2025-fact,delbrouck-etal-2022-improving,10.5555/3618408.3619699}. For example, in response to the query “Is lemon effective for curing cancer?”, a standard retriever may prioritize passages containing lexical associations with “lemon,” “cancer,” or even misinformation, while failing to surface clinically substantiated evidence. An evidence-based retriever, by contrast, must elevate passages reporting verified trial outcomes or medical consensus. When generation models are conditioned on retrieval distributions lacking evidential precision, the likelihood of propagating incomplete, biased, or spurious information increases substantially, amplifying hallucinations and degrading factual reliability \cite{hu-etal-2025-removal}.

\begin{figure*}[!ht]
    \centering
   \includegraphics[width=0.95\textwidth]{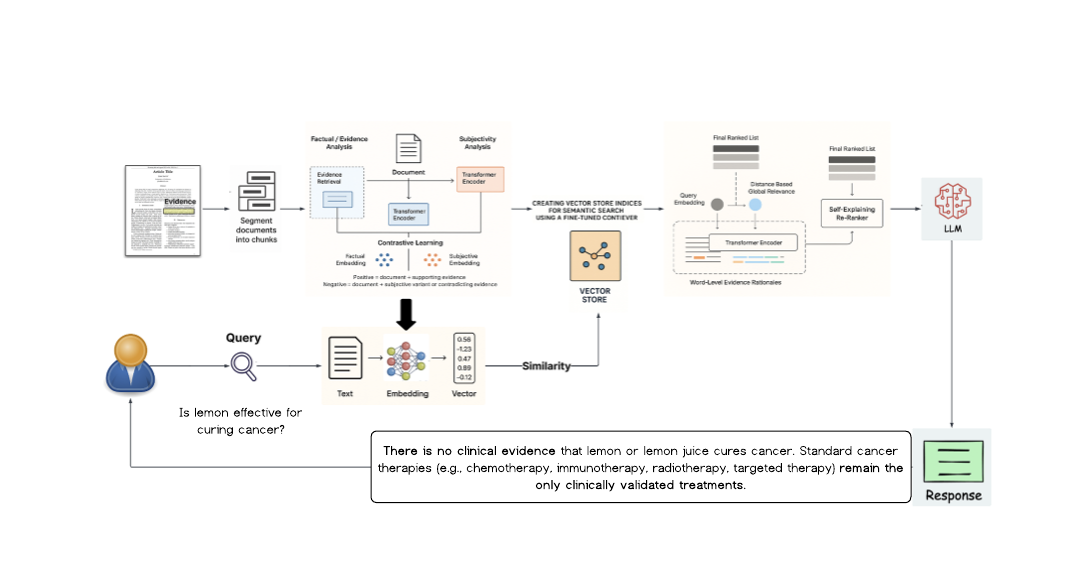}
    \caption{The pipeline for self-Explaining Re-Ranking with Contrastive Evidence Selection for Retrieval-Augmented Generation.}  
    \label{fig:cer}
\end{figure*}

Recent work has attempted to address these deficiencies through contrastive learning \cite{izacard2022unsuperviseddenseinformationretrieval}, training retrievers to discriminate between relevant and irrelevant passages by maximizing representational separation between positive and negative samples \cite{wu-etal-2025-medical,sun-etal-2025-fact,sriram-etal-2024-contrastive}. However, these approaches typically treat all negative instances as homogeneous, disregarding the various semantic functions they may serve with respect to a target claim. For example, providing corroborating evidence, presenting contradictory findings, or being entirely unrelated. Moreover, current contrastive retrievers optimize relevance implicitly through similarity scores, offering limited transparency regarding the retrieval rationale and leaving the underlying evidential reasoning process opaque.

To fill these limitations, we propose Self-Explaining Contrastive Evidence Re-ranking (CER), which integrates triplet-based contrastive learning with explicit evidential attribution. CER fine-tunes Contriever \cite{izacard2022unsuperviseddenseinformationretrieval} using a cosine and euclidean triplet loss to restructure the embedding space, as factual evidence is systematically drawn closer to the query representation, while subjective, contradictory, or irrelevant content is repelled. Empirical evaluation on large-scale clinical trial corpora demonstrates significant gains in retrieval accuracy. %, indicating that CER yields a more precise, interpretable, and evidence-aligned retrieval distribution for RAG systems. %Although assessed in the biomedical domain, CER is broadly generalizable and applicable to any setting in which factual rigor and interpretability are essential requirements.

%------------------------------------------------------
\section{Related Work} \label{sec:firstpage}
%------------------------------------------------------
 \citet{chatzikyriakidis-natsina-2025-poetry} generate Modern Greek interwar poetry with GPT-4-turbo and GPT-4o using RAG and contrastive learning. RAG retrieves thematically and stylistically similar poems to guide generation, while the contrastive variant adds opposing examples for style control. Expert evaluations and quantitative metrics (vocabulary density, average words per sentence, readability index) show that RAG improves style and thematic consistency. \citet{sriram-etal-2024-contrastive} create a dense retriever fine-tuned with contrastive learning to improve evidence retrieval for complex fact-checking. Contrastive Fact-Checking Reranker (CFR), extends Contriever \cite{izacard2022unsuperviseddenseinformationretrieval} using multiple supervision signals: GPT-4 distillation, LERC-based answer equivalence, and human-annotated gold data from the AVeriTeC dataset. The model enhances second-stage retrieval by ranking documents that better support claim verification. Experiments show a 6\% improvement in veracity classification accuracy and 9\% higher top-document relevance on AVeriTeC, with consistent gains across benchmarks, demonstrating improved reasoning and robustness in real-world fact-checking.

\section{The Proposed Method}
The pipeline is decomposed into two complementary stages: \textbf{contrastive retrieval}, which learns an evidence-sensitive embedding space through (i) document chunking and contrastive training and (ii) contextual retrieval; and \textbf{self-explaining re-ranking}, which provides fine-grained attribution signals to identify the most query-aligned passages through (iii) self-explaining re-ranking. The selected evidence—augmented with token-level rationales—is then passed to an LLM, which generates a response grounded in verifiable evidence. The system subsequently returns the final answer to the user. In this design, the first stage ensures that retrieval prioritizes factual evidence rather than merely topical content, while the second stage refines ranking via token-level semantic contributions, enabling interpretable and evidence-based selection of supporting passages. Next, we describe the method in detail.

\section{Evaluation and Initial Results}
We evaluated only the proposed fine-tuned Contriver model, reporting results using recall@K and precision@K (e.g., K=5).

\begin{figure}[!htb]
    \centering
   \includegraphics[width=0.23\textwidth]{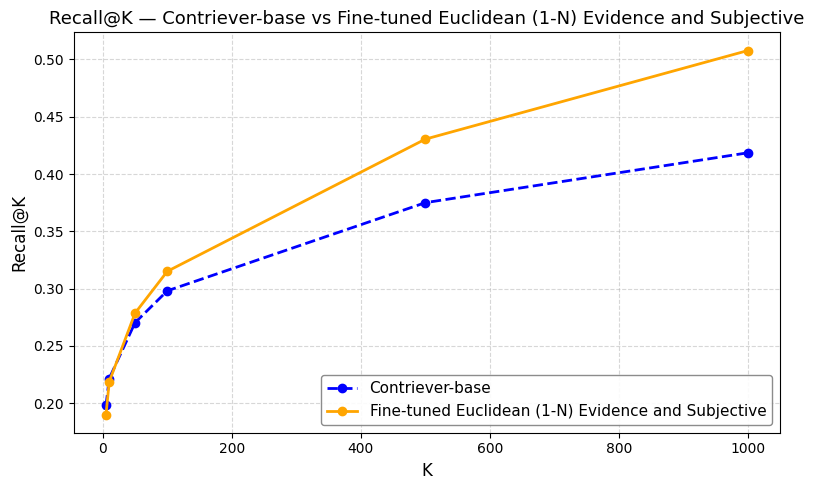}
    \includegraphics[width=0.23\textwidth]{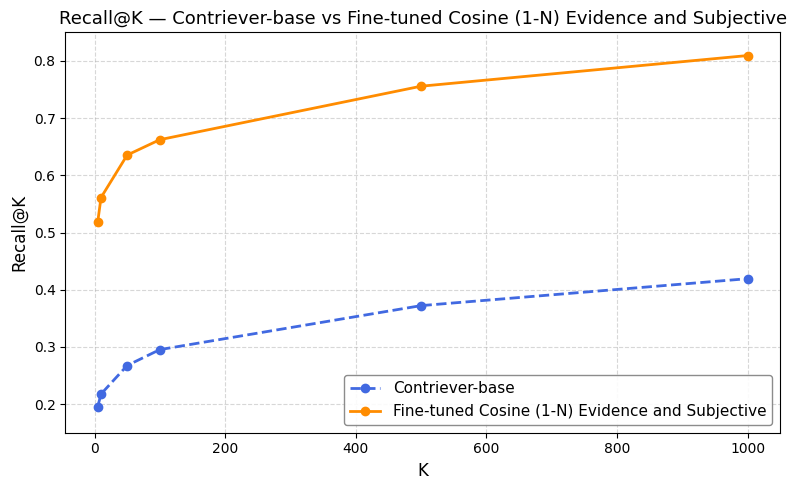}
    \caption{Self-learning and fine-tuned models evaluated using both Euclidean and cosine distance metrics.}  
    \label{fig:results0}
\end{figure}

\begin{figure}[!htb]
    \centering
   \includegraphics[width=0.21\textwidth]{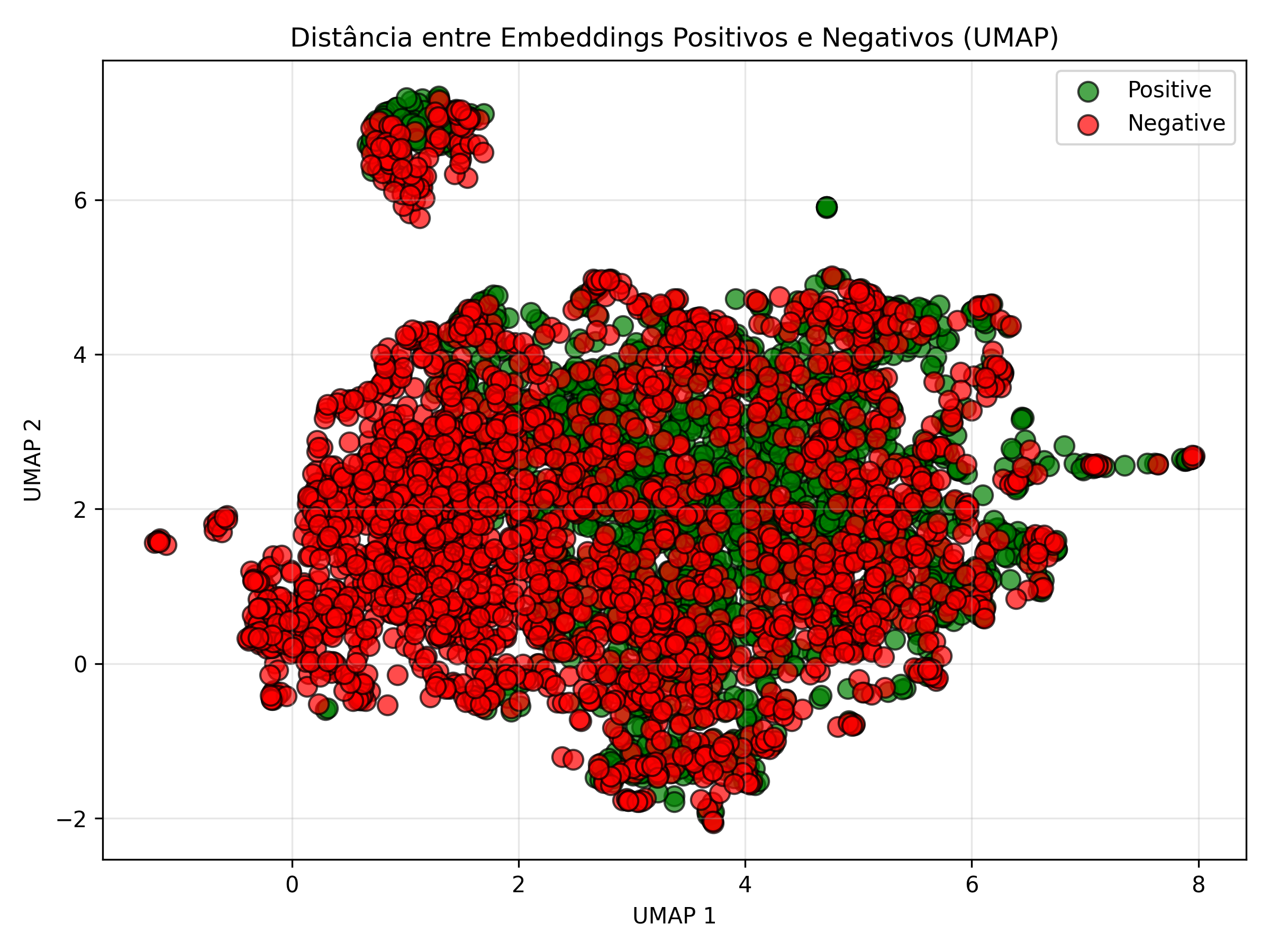}
    \includegraphics[width=0.21\textwidth]{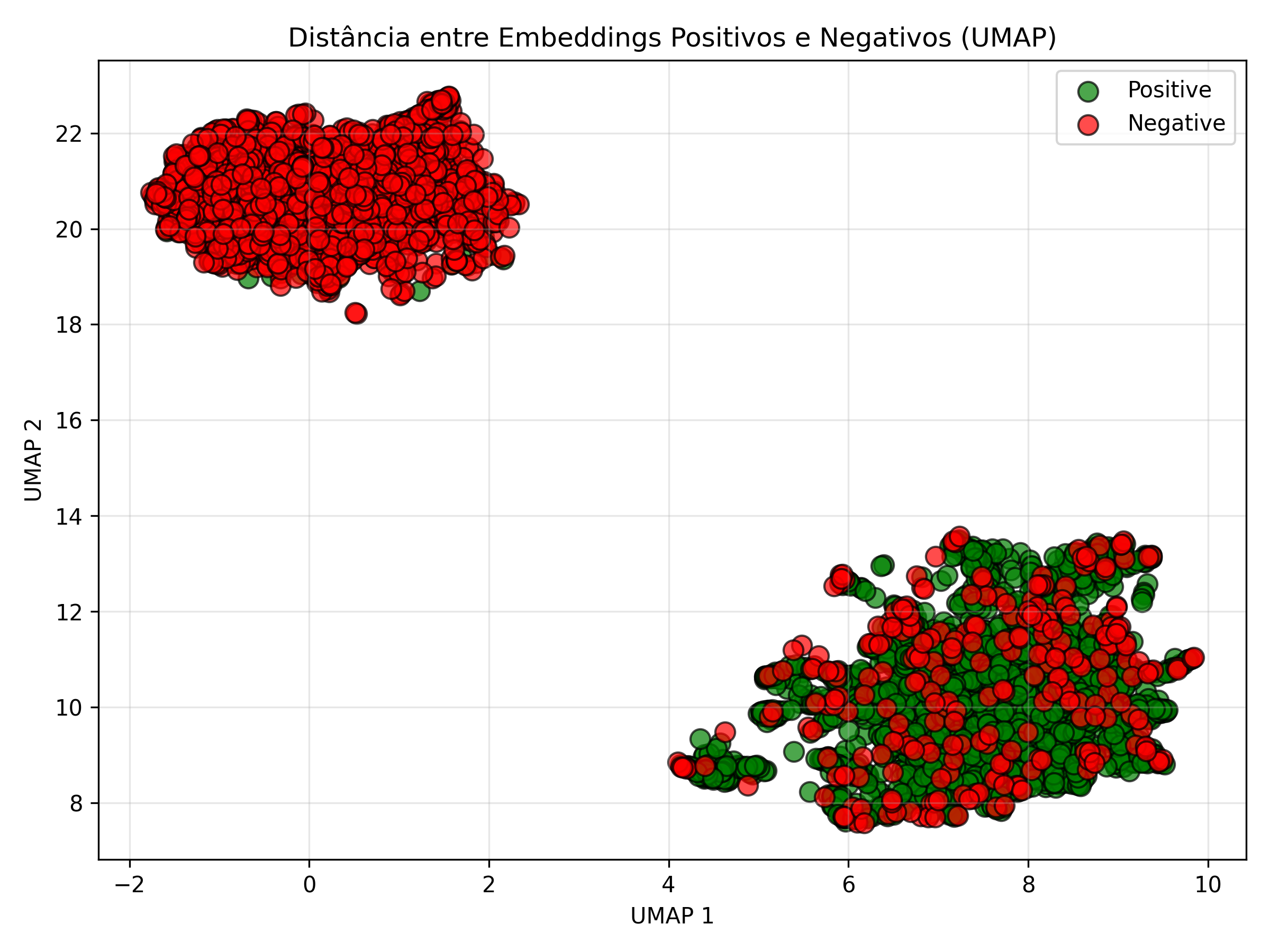}
    \caption{Results for the Contriever self-learning model (left) and the Contriever fine-tuned model (right). For the self-learning model, the average distance among positive pairs is Intra-Pos = 0.5716 and among negative pairs is Intra-Neg = 0.5977; the average distance between positives and negatives is Inter = 0.5901. For the fine-tuned model, the average distance among positive pairs is Intra-Pos = 0.7766 and among negative pairs is Intra-Neg = 0.8141; the average distance between positives and negatives is Inter = 0.8110.}  
    \label{fig:results1}
\end{figure}

As shown in \ref{fig:results1}, The UMAP projections show that our fine-tuned model, trained with the proposed subjectivity-based triplet selection, produces a much clearer separation between evidential (positive) and non-evidential (negative) embeddings. While the self-learning baseline displays substantial overlap between classes, the fine-tuned model forms well-separated clusters, indicating that contrastive training effectively distinguishes evidence from non-evidence. We also evaluated the efficiency between applying Euclidean and cosine distance metrics, as shown in Figure \ref{fig:results0}.

%\section{Final Remarks}
%This thesis highlights the necessity of integrating explainability into NLP models to ensure the ethical deployment of AI, focusing on critical tasks such as automated fact-checking and hate speech detection. By proposing benchmark datasets and novel methods and systems, it lays the foundation for more transparent and socially responsible AI solutions in misinformation and hate speech detection. Future research will explore the generalization of these techniques to other languages, tasks, and deep learning architectures, further advancing the goal of responsible AI. 

\newpage
\section*{Acknowledgements}
The authors are grateful to São Paulo Research Foundation - FAPESP (grants \#2025/01118-2 and \#2024/04890-5) for financial support.

\section*{Limitations}
These findings represent only initial results obtained from experiments using the fine-tuned model trained with our proposed criteria for selecting positive and negative examples in the triplet loss for contrastive learning. While the results indicate that the selection strategy is effective in structuring the embedding space and separating evidential from non-evidential content, they should be interpreted as preliminary evidence rather than definitive validation. More robust experiments are still required to thoroughly evaluate the proposed method, both in the contrastive retrieval stage and in the re-ranking stage. Furthermore, no assessment has yet been conducted on the quality of the explanations generated by the self-explaining component, which remains an essential step for understanding the interpretability and reliability of the approach.
%--------------------------------------------
\bibliographystyle{ACM-Reference-Format}
\bibliography{sample-base.bib}

%--------------------------------------------

\end{document}